\documentclass[letterpaper, 10 pt, conference]{ieeeconf}  

\IEEEoverridecommandlockouts                              
\makeatletter

\newcommand{\Rmnum}[1]{\expandafter\@slowromancap\romannumeral #1@}
\makeatother

\usepackage[utf8]{inputenc}
\usepackage[T1]{fontenc}
\usepackage{cite}
\usepackage{graphicx} 
\usepackage{tikz}
\usetikzlibrary{positioning}
\usepackage{subfig}

\usepackage{amsmath}
\usepackage{amsfonts}
\usepackage{amssymb}
\usepackage{siunitx}
\usepackage{utfsym}
\usepackage{hyperref}
\usepackage{cleveref}

\title{\LARGE \bf
100 Drivers, 2200 km: A Natural Dataset of Driving Style toward Human-centered Intelligent Driving Systems
}

\author{Chaopeng Zhang$^1$, Wenshuo Wang$^{1*}$, Zhaokun Chen$^1$, ~Junqiang Xi$^{1*}$ 
\thanks{*This work was supported by the National Natural Science Foundation of China (Grant No.: 52272411).}
\thanks{$^1$C. Zhang, W. Wang, J. Xi, and Z. Chen are with the School of Mechanical Engineering, Beijing Institute of Technology, Beijing, China 100081.
         {\tt\small cpzhang@bit.edu.cn; ws.wang@bit.edu.cn}}
}

\begin{document}

\maketitle
\thispagestyle{empty}
\pagestyle{empty}

\begin{abstract}
Effective driving style analysis is critical to developing human-centered intelligent driving systems that consider drivers' preferences. However, the approaches and conclusions of most related studies are diverse and inconsistent because no unified datasets tagged with driving styles exist as a reliable benchmark. The absence of explicit driving style labels makes verifying different approaches and algorithms difficult. This paper provides a new benchmark by constructing a natural dataset of Driving Style (100-DrivingStyle) tagged with the subjective evaluation of 100 drivers' driving styles. In this dataset, the subjective quantification of each driver's driving style is from themselves and an expert according to the Likert-scale questionnaire. The testing routes are selected to cover various driving scenarios, including highways, urban, highway ramps, and signalized traffic. The collected driving data consists of lateral and longitudinal manipulation information, including steering angle, steering speed, lateral acceleration, throttle position, throttle rate, brake pressure, etc. This dataset is the first to provide detailed manipulation data with driving-style tags, and we demonstrate its benchmark function using six classifiers. The 100-DrivingStyle dataset is available via https://github.com/chaopengzhang/100-DrivingStyle-Dataset
\end{abstract}

\section{Introduction}
Autonomous vehicles are rapidly promoted due to emerging perception, decision-making, and control technologies. Still, it is not so fast to safely bring them on public roads due to many technical and social challenges that require enormous effort \cite{wang2022social}. Therefore, on-road vehicles will be in the form of human-machine co-driving systems  for the foreseeable future \cite{russell2016motor}. However, human drivers have diverse and time-varying individual driving preferences, which makes it challenging to design a human-centered intelligent driving system. For example, some drivers seek sensations and thrill \cite{french1993decision}, while others pursue comfort during driving. Many researchers developed data-driven approaches with advanced machine learning techniques for driving style analysis, clustering, and recognition to capture the driving preference of human drivers in various driving scenarios\cite{chu2020self,zhang2022generative,li2018estimating,barendswaard2019classification,deng2019curve,hu2022study}. These data-driven approaches usually rely on high-quality driving data across drivers. However, no publicly available datasets specifically designed for driving style analysis exist. Some works use public trajectory datasets such as SPMD\cite{bezzina2014safety}, and NGSIM \cite{alexiadis2004next} to analyze driving style (e.g., clustering) but lack ground truth for evaluating algorithm performance. That is, these public datasets do not have driving style labels, making it impossible to verify the subjective and objective consistency of the algorithm performance. Some works also collected data from the lab's driving simulators \cite{wang2017driving}, but the data fidelity is far from the driving behavior in real-world scenarios. In addition to human personality, the driving style can be influenced by driving routes and traffic conditions. For example, the speed on a free highway is usually much faster than the speed on congested urban roads.

To address the above issues, we constructed a new naturalistic driving-style dataset of 100 drivers from a well-chosen and fixed driving route that covers diverse driving scenarios. The main goal of our driving-style dataset is to collect driving behavior data with driving-style tags as a benchmark. We also exclude the influence of routes and vehicle types on driving styles by using one vehicle with data acquisition systems and fixing a driving route to collect data. To obtain drivers' subjective evaluation of driving style, we used a five-point Likert scale to evaluate driving style comprehensively. The 100-DrivingStyle dataset consists of complete manipulation data from (controller area network (CAN) signals and subjective evaluation of driving style for their driving behavior, providing solid data support for driving style research.

\section{Related Works}
In recent years, many public datasets have facilitated the research on human-centered intelligent driving systems. Still, they rarely contain driver IDs with tags of driving styles \textit{or} only record a short period of driving behaviors for individual drivers. The small data on individuals' driving behaviors can not reflect the driver's driving styles. Since the 100-DrivingStyle dataset is primarily for driving style analysis, we here do not review the existing datasets unrelated to driving style. Table \ref{tab: Comparison of dataset for driving style} compares multiple datasets in detail, which contain long driving data of different drivers. 

\begin{table*}[]
\centering
\caption{Comparison of The Existing Naturalistic Driving Datasets}
\label{tab: Comparison of dataset for driving style}
\vspace{0ex}
\begin{tabular}{lllll}
\hline\hline
                             & 100-DrivingStyle                 & SPMD      & NGSIM     & high-D   \\ \hline
driving style labels  & \usym{2713}     & \usym{2717}    & \usym{2717}  & \usym{2717}   \\
driver information  & \usym{2713}     & \usym{2717}    & \usym{2717}  & \usym{2717}   \\
fixed vehicle  & \usym{2713}     & \usym{2717}    & \usym{2717}  & \usym{2717}   \\
fixed route     & \usym{2713}    & \usym{2717}  & \usym{2713} & \usym{2713}    \\
Operation data    & \usym{2713} & \usym{2713} &\usym{2717}    & \usym{2717} \\
avg. driving time per driver & $\sim$ 30 min    &    -  & < 1 min      & < 1 min \\
record time       & 50 h       &   -   & 2.5h   & 16.5h  \\
driving scene     & \begin{tabular}[c]{@{}l@{}}urban, highway,curves,\\ urban intersections,\\ roundabouts, ramps\end{tabular} & \begin{tabular}[c]{@{}l@{}}urban, highway,\\ urban intersections,\\ roundabouts, ramps\end{tabular} & \begin{tabular}[c]{@{}l@{}}urban, highway,\\ urban intersections,\end{tabular} & highway    \\

record frequency   & 100 Hz   & 10 Hz    & 10 Hz    & 25 Hz     \\
sensors       & on-board    & on-board       & stationary camera      & stationary drone 
\\ \hline\hline       
\end{tabular}
\end{table*}


\subsection{On-Board Sensor Data} A variety of datasets from vehicles with on-board sensors have been released. One main advantage of these datasets is that the operation information can be accessed, such as steering angle and gas/brake pedals. For instance, the SPMD database recorded the naturalistic driving of 2842 equipment vehicles (e.g., passenger vehicles, truck fleets, and transit buses) in Ann Arbor, Michigan, from 2012 to 2014. Specifically, the SPMD database provides the proceeding vehicle's information (e.g., relative distance and relative speed) and the subjective vehicle information (e.g., speed, steering, and acceleration/brake pedal position). However, the SPMD database is not designed for driving style analysis, and \textit{does not contain a driving-style label for individual drivers}. In addition, the driving route of each driver is different in the SPMD. Therefore, the influence of different routes on driving style cannot be evaluated correctly.

\subsection{Stationary Sensor Data} In recent years, stationary or drone cameras have been set up to provide high-quality bird-view recordings of videos. 
For example, the NGSIM dataset is constructed using cameras mounted on the top of buildings at four different places, covering a region of interest (RoI) with a length of about 500 to 640 meters. The NGSIM dataset can provide the speed and acceleration of vehicles falling into RoI. Another well-known dataset, High-D \cite{krajewski2018highd}, recorded natural driving trajectories at German highways via drones. HighD provides data such as relative position, velocity, acceleration, heading angle, time to collisions, lane ID, etc. The length of the recorded road segment covers about 400 \si{m}. HighD provides data of a recording duration of about 16.5 hours in total, but \textit{the average driving time per driver/vehicle is only tens of seconds}. The common disadvantages of this method are 
\begin{itemize}
\item \textbf{Short recordings:} the driving time of a single driver is too short to be enough to analyze driving styles; and
\item \textbf{Single scenarios:} the diversity of driving scenarios is not rich enough to ensure the generalization ability of driving style analysis.
\end{itemize}

\subsection{Summary and Contribution}
To address the limitations of on-board sensors and fixed-camera sensors, we constructed a new naturalistic driving dataset for driving styles with the following contributions:
\begin{itemize}
    \item \textbf{Driving style tags:} Each driver's subjective driving style is provided as a valid benchmark for driving style analysis and algorithm development.
    \item \textbf{Long recording of individual drivers:} 100 drivers with different driving experiences were covered, with the average driving time per driver being more than 30 minutes, ensuring the sufficiency and diversity of the dataset.
    \item \textbf{Unified testing route:} The driving route is unified to avoid the influence of different routes on driving style.
    \item \textbf{Diversity in scenarios:} The driving route covers daily driving scenarios, including urbans, highways, signalized intersections, ramps, roundabouts, curves, etc.
    \item \textbf{Operation information:} Complete manipulation data was collected, such as vehicle speed, gas pedal, opening rate of gas pedal, steering wheel angle, steering wheel speed, etc., which can promote the algorithm development of driving styles to the control layer.
\end{itemize}

\section{The Driving-Style Dataset Collection}
This section will introduce the data collection preparation (participants, equipment, route selection), experiment procedure, and subjective quantification of driving style. 
\subsection{Participants}
Driving styles can be influenced by personalities, such as age, gender, occupation (e.g., taxi driver), driving experience, etc. To cover a variety of drivers, we recruited 100 drivers (83 males and 17 females) with diverse jobs, ages, and driving experience. Table \ref{tab: Statistical summary of tested drivers} lists the statistical information of these drivers. Data collection for the 100 drivers runs from February 2022 to October 2022.

\begin{table}[t]
\centering
\caption{The Summary of the 100 Drivers.}
\label{tab: Statistical summary of tested drivers}
\begin{tabular}{ll|ll|ll}
\hline\hline
\multicolumn{2}{l|}{Age} & \multicolumn{2}{l|}{Driver experience} & \multicolumn{2}{l}{Occupation} \\
\hline
Range      & Number      & Range             & Number             & Occupation        & Number     \\
\hline
21-25      & 3           & 1-5               & 12                 & bus driver        & 38         \\
26-30      & 8           & 6-10              & 20                 & taxi driver       & 10         \\
31-35      & 14          & 11-15             & 21                 & ride-hailing      & 9          \\
36-40      & 21          & 16-20             & 20                 & self-employed     & 9          \\
41-45      & 18          & 21-25             & 12                 & engineer          & 4          \\
46-50      & 16          & 26-30             & 8                  & worker            & 4          \\
51-60      & 17          & 31-35             & 7                  & teacher           & 3          \\
56-60      & 3           &-                    &-                 & freelancer        & 16         \\- 
           &-              &-                    & -                    & others            & 7         \\
\hline\hline
\end{tabular}
\end{table}

\subsection{Hardware and Software Setup}
\begin{figure}[t]
    \centering
    \includegraphics[width=\linewidth]{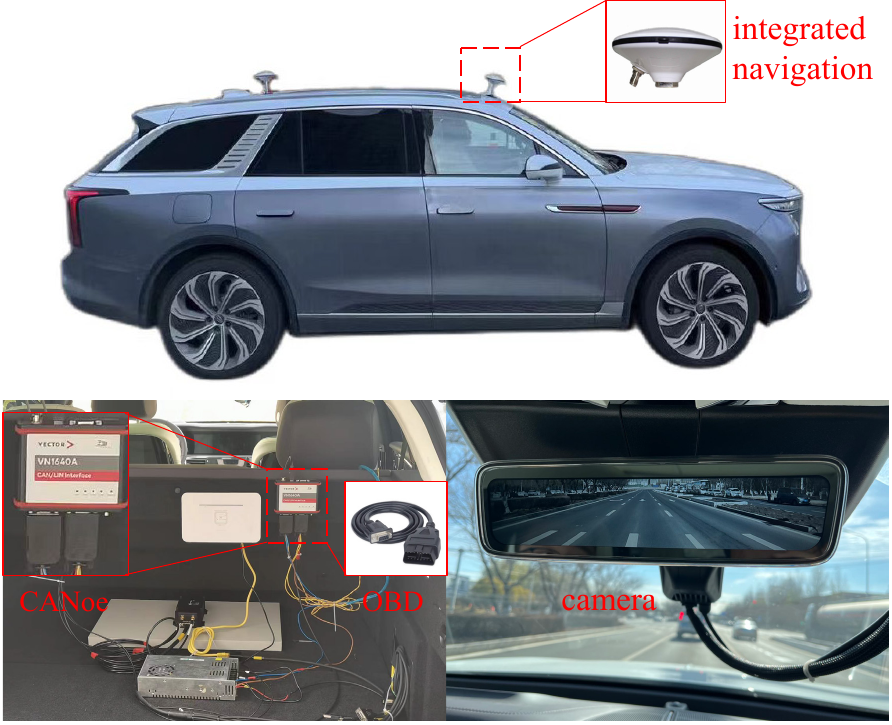}
    \caption{Architecture of data-acquisition system}
    \label{fig: Equipment}
\end{figure}

Fig.~\ref{fig: Equipment} shows our testing vehicle equipped with a data-acquisition system. The data-acquisition system mainly contains a vehicle CAN network, an integrated navigation system, and one front-view camera. The CAN information is transmitted through the on board diagnostics (OBD) interface and decoded using the  CAN database file provided by the vehicle manufacturer. Table \ref{table: recorded data} lists the collected CAN Information. The vehicle's integrated navigation system is installed to obtain positioning and orientation information. The visual perception system collects traffic information, such as traffic lights, lane lines, etc.
CANoe is utilized to synchronize and record all real-time acquisitions at a rate of 100 Hz. The CAN network load rate does not exceed 30\% to ensure the stability of data transmission. 
Data acquisition software runs on a laptop with an Intel Core i7 processor at 4.1GHz and 16GB RAM.

\begin{table}
\centering
\caption{Recorded Data Information}
\label{table: recorded data}
\begin{tabular}{llll}
\hline\hline
Features & Definition & Unit & Precision\\
\hline
$v$ & longitudinal velocity & \si{km/h} & 0.01 \\
$a_{x}$, $a_{y}$ & lateral and longitudinal acceleration & \si{m/s}$^{2}$ & 0.03\\
$\beta$ & throttle opening & \% & 0.4\\
$\dot{\beta}$ & derivative of throttle opening & \%/s & 2\\
$\delta$ & steering angle & \si{\degree} & 0.1\\
$\dot{\delta}$ & derivative of steering angle & \si{\degree/s} & 0.5\\
$\psi$ & yaw rate & deg/s & 0.05\\
$b$ & braking pressure & bar & 0.1\\
\hline\hline
\end{tabular}
\end{table}

\subsection{Driving Routes}
\begin{figure}[t]
    \centering
    \includegraphics[width=\linewidth]{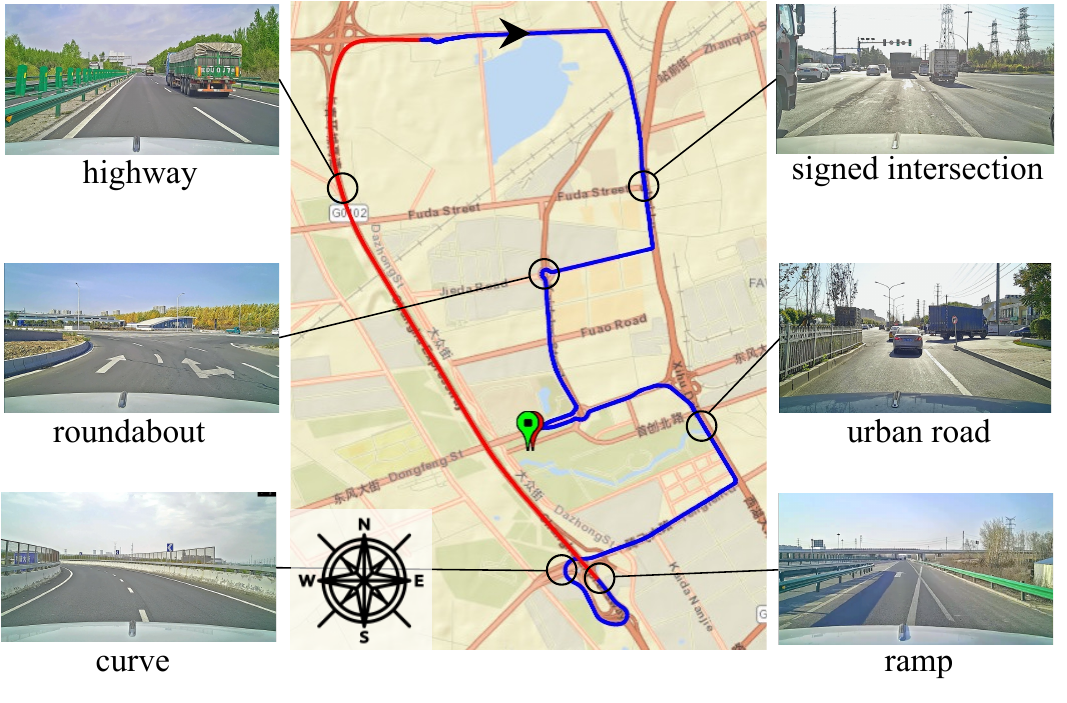}
    \caption{Driving routes for data collection.}
    \label{fig: Driving routes}
\end{figure}

To cover as many daily driving scenarios as possible, we selected driving routes in Changchun, China, as shown in Fig.~\ref{fig: Driving routes}. The driving route consists of 14.4 km of urban roads and 7.6 km of highways,  covering signed intersections, roundabouts, ramps, curves, and so on. 

\subsection{Data Collection Procedure}
\begin{figure}[t]
    \centering
    \includegraphics[width=0.8\linewidth]{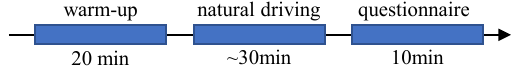}
    \caption{Data collection pipeline}
    \label{fig: Data collection pipeline}
\end{figure}

As shown in Fig.~\ref{fig: Data collection pipeline}, a single participant takes about one hour to collect data, including a warm-up session, natural driving session, and questionnaire session. To familiarize the participants with the vehicle's operation and interaction, we prepared a 20-minute warm-up session. During warm-up sessions, we explained safety issues to the driver, clarified any doubts, and informed him/her to drive according to his/her  own style. The driver previewed the navigation route and test drives the vehicle. 

After the warm-up session, he/she drived naturally according to their driving style by following the selected navigation route. Running traffic lights while driving is prohibited, but speeding is not. It is up to the driver to decide whether to exceed the speed limit or not. Because it is common that some drivers often exceed the speed limit in their daily driving when there is no traffic monitoring. We believe that this is a reflection of the driver's driving style, albeit irrational. During the driving, once the data-collection session is triggered, our data-acquisition system automatically record the participant's driving operations. As drivers often have different driving styles in the city and on the highway, we stoped at highway toll booths for 3 minutes to allow drivers to adjust to the change in environment.

After the driving session, we immediately asked the driver participant to complete the driving style questionnaire, providing their age, gender, occupation, driving experience, and self-evaluation of driving styles. However, it is difficult to get drivers to pinpoint their driving styles because drivers have varying levels of understanding. Therefore, we designed the driving style questionnaire based on the Likert scale \cite{joshi2015likert,padilla2020adaptation}, an easy-to-answer questionnaire paradigm. Drivers did not have to answer exactly what type of driving style they belong to, but simply indicated their aggressiveness level compared to other drivers. Researchers were on hand to clarify the driver's doubts about the questionnaire but do not interfere with the driver's subjective judgment. Meanwhile, we also invited an experienced expert in the car to score the participant's driving style subjectively. 

\subsection{Driving Style Questionnaire}
Each driver participant and expert reported their feelings of aggressive levels independently, and the aggressive level was quantified into five scales. We considered three factors (safety, risk-taking, and stimulation-seeking) to design the five levels of aggressiveness.
\begin{itemize}
    \item Level 1 (very poor aggressive): The driver strictly abides by safe driving regulations, averse to risk and stimulation.
    \item Level 2 (poor aggressive): The driver abides by safe driving regulations and tries to avoid risks.
    \item Level 3 (a bit aggressive): The driver abides by safe driving regulations but occasionally pursues stimulation under the premise of safety.
    \item Level 4 (aggressive): The driver occasionally violates safe driving regulations, seeks stimulation, and can tolerate certain risks.
    \item Level 5 (very aggressive): The driver often violates safe driving regulations, pursues stimulation, and is willing to take higher risks.
\end{itemize}


\begin{figure}
    \centering
    \includegraphics[scale=0.85]{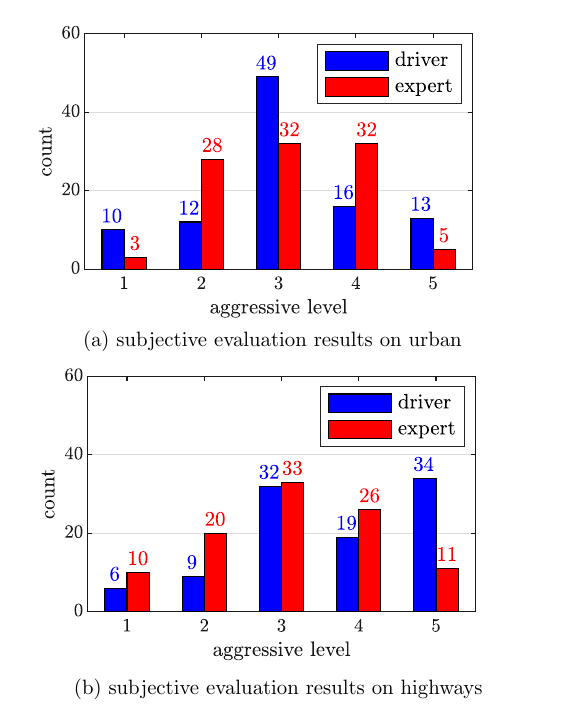}
    \caption{Subjective evaluation results of the drivers and expert on different traffic environments. (a) urban (b) highway}
    \label{fig: Subjective driving style of the drivers and expert}
\end{figure}

\section{Dataset Analysis}
\subsection{Subjective Driving Style}
We subjectively evaluate individuals' driving styles according to the participant's self-evaluation and expert evaluation, providing a reference or ground truth. Fig. \ref{fig: Subjective driving style of the drivers and expert} shows subjective evaluation results of driving styles on different traffic environments of urban and highways. On urban roads, most drivers ($\sim$ 50\%) drive cars in a little bit aggressive style (level 3) which may be attributed to the subjective cognition of individual drivers. Individual drivers tend to think they are normal and complain that others are too aggressive or conservative. The literature \cite{tillmann1949accident} came to similar conclusions by comparing 96 high-accident drivers with 100 accident-free drivers. Aggressive drivers usually do not realize that they are aggressive and often criticize their own driving  mistakes in other driver. However, most drivers raised their aggressive level in evaluating driving style on the highway because they find themselves driving more aggressively on the highway than in the city. This phenomenon proves that traffic environments can influence the driver's driving style. Coincidentally, the expert's evaluation of driving style obeys the normal distribution approximately. In addition, experts have rich experience and evaluate drivers' driving styles with an inherent unified standard. The experts' evaluations showed little difference between the city and the highway because the experts compared the driving behaviors of the drivers from a horizontal comparison point of view. Driving style evaluation will be published with the dataset. 

\begin{figure}[t]
    \centering\includegraphics[width=0.8\linewidth]{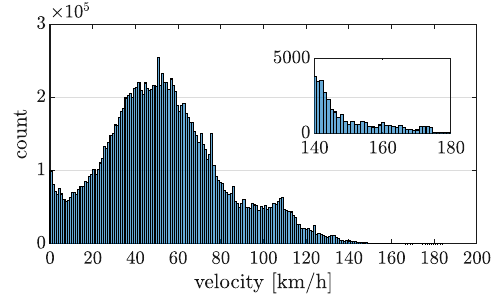}
    \caption{Histogram of velocity in the 100-DrivingStyle dataset}
    \label{fig: Histogram of velocity in NDSD}
\end{figure}

\subsection{Statistical Evaluation}
The dataset contains 100 records, each with a duration of 25$\sim$35 minutes, for a total of about 3000 minutes. The recording time covers the morning and afternoon from February 15, 2022, to October 15, 2022. The weather covers sunny, cloudy, rainy, and snowy days during this period. Because the route contains both urban and highway, the speeds cover a wide range, as shown in Fig. \ref{fig: Histogram of velocity in NDSD}. Three distinct peaks for the most frequent velocities can be observed. The lowest velocity peak is near 0 km/h, corresponding to the start-stop at red lights and traffic jams. The peaks near 55 km/h and 110 km/h are typical speeds for drivers. Velocity peaks close to 55 km/h and 110 km/h correspond to the typical speeds of drivers in the city and highway, respectively. Furthermore, the typical speeds do not exceed the local traffic speed limits, which are 60 km/h for urban roads and 120 km/h for highways.

\section{Driving Style Recognition Verification}
\label{subsection: Driving Style Recognition}

To illustrate the benchmark function of the proposed 100-DrivingStyle dataset, we developed six driving style classifiers with urban driving data. The procedure includes feature selection, common factors extraction, driving style classifier development and verification. 
Following the above process, we have developed six driving style classifiers based on  Support Vector Machine (SVM), Linear Discriminant Analysis (LDA), Naive Bayes Classifier (NBC), ensemble classifier, K-Nearest Neighbor (KNN), and decision tree.

\subsection{Feature Selection and Data Processing}
To comprehensively describe the longitudinal and lateral driving characteristics of the drivers, the following variables were chosen. 
\begin{enumerate}
    \item [1)] The vehicle speed $v$, which reflects individual speed preference and risk tolerance.
    \item [2)] The longitudinal acceleration $a_x^+$ and deceleration $a_x^-$ which reflect the driver's preferred acceleration and deceleration.
    \item [3)] The vehicle yaw rate $\psi$, which can reflect driver's steering habit.
    \item [4)] The vehicle lateral acceleration $a_y$, which reflects driver's side slip risk tolerance.
    \item [5)] The throttle opening $\alpha$, which reflects the driver's driving intent and accelerator characteristics.
\end{enumerate}

In order to describe the driving characteristic, we calculated the statistics ( maximum, mean, standard deviation) of driving data, and the driving data of the $i-th$ driver can be described as $\mathbf{x}_i \in \mathbb{R}^p$, and $p$ denotes the dimensions of driving data. 

To assign moderate subjective driving style labels to the $100$ drivers as ground truth $(s_\mathrm{sub})$, we average the scores given by experts and drivers, followed by rounding the weighted scores to the nearest whole number. For instance, if a driver's scores are levels $1$ and $2$, the weighted score after averaging would be level $2$. The subjective driving style labels of the $100$ drivers are demonstrated in Table \ref{table: The Subjective Driving Style Labels}. It should be noted that the count of level 1 is $0$, attributable to the non-existence of samples where both driver and expert ratings simultaneously equal one. The labels are used to train and verify  driving style classifiers.

\begin{table}[]
\centering
\caption{The Subjective Driving Style Labels $(s_\mathrm{sub})$ of The $100$ Drivers}
\label{table: The Subjective Driving Style Labels}
\begin{tabular}{llllll}
\hline\hline
Levels &Level 1 &Level 2  &Level 3  &Level 4  &Level 5 \\
\hline
Number                     & 0 & 12 & 46 & 35 & 7 \\
\hline\hline
\end{tabular}
\end{table}

\subsection{Factor Analysis}
The driving data statistics $\mathbf{y}$ are usually in a high dimension, preventing driving style analysis and interpretation. To solve this problem, we introduce factor analysis to transfer the high-dimension driving data into a low-dimensional space for efficient analysis by discovering some common factors influencing driving behavior statistics. \cite{yong2013beginner}. 

Given each driver's driving statistics $\{\mathbf{x}_{i}\}_{i=1}^{N}$, we stack them sequentially and obtain a matrix $\mathbf{X}$ of which each row represents the statistic feature as $\mathbf{X} = \left[\mathbf{x}_1, \mathbf{x}_2, \hdots, \mathbf{x}_N \right]^\top $, where $N=100$ denotes the total number of drivers.
Using factor analysis, all driving data $\mathbf{X}$ can be described in the common factor space as,
\begin{equation}
\mathbf{Y} = 
\begin{bmatrix}
\boldsymbol{y}_{1}\\
\boldsymbol{y}_{2} \\
\vdots\\
\boldsymbol{y}_{m}
\end{bmatrix}
=
\begin{bmatrix}
y_{11} & y_{12} & \dots & y_{1N} \\
y_{21} & y_{22} & \dots & y_{2N} \\
\vdots & \vdots & \ddots & \vdots \\
y_{m1} & y_{m2} & \dots & y_{mN} \\
\end{bmatrix}
\end{equation}
where $m\ll p$ denotes the dimensions of common factors. The value is determined according to Kaiser's criterion\cite{kaiser1960application}, and we get three common factors $(m=3)$ that account for $83.3$\% of the variance in the urban data. 

\subsection{Benchmark Verification of Driving Style Classifiers}
\begin{figure}[h]
    \centering
    \includegraphics[width=\linewidth]{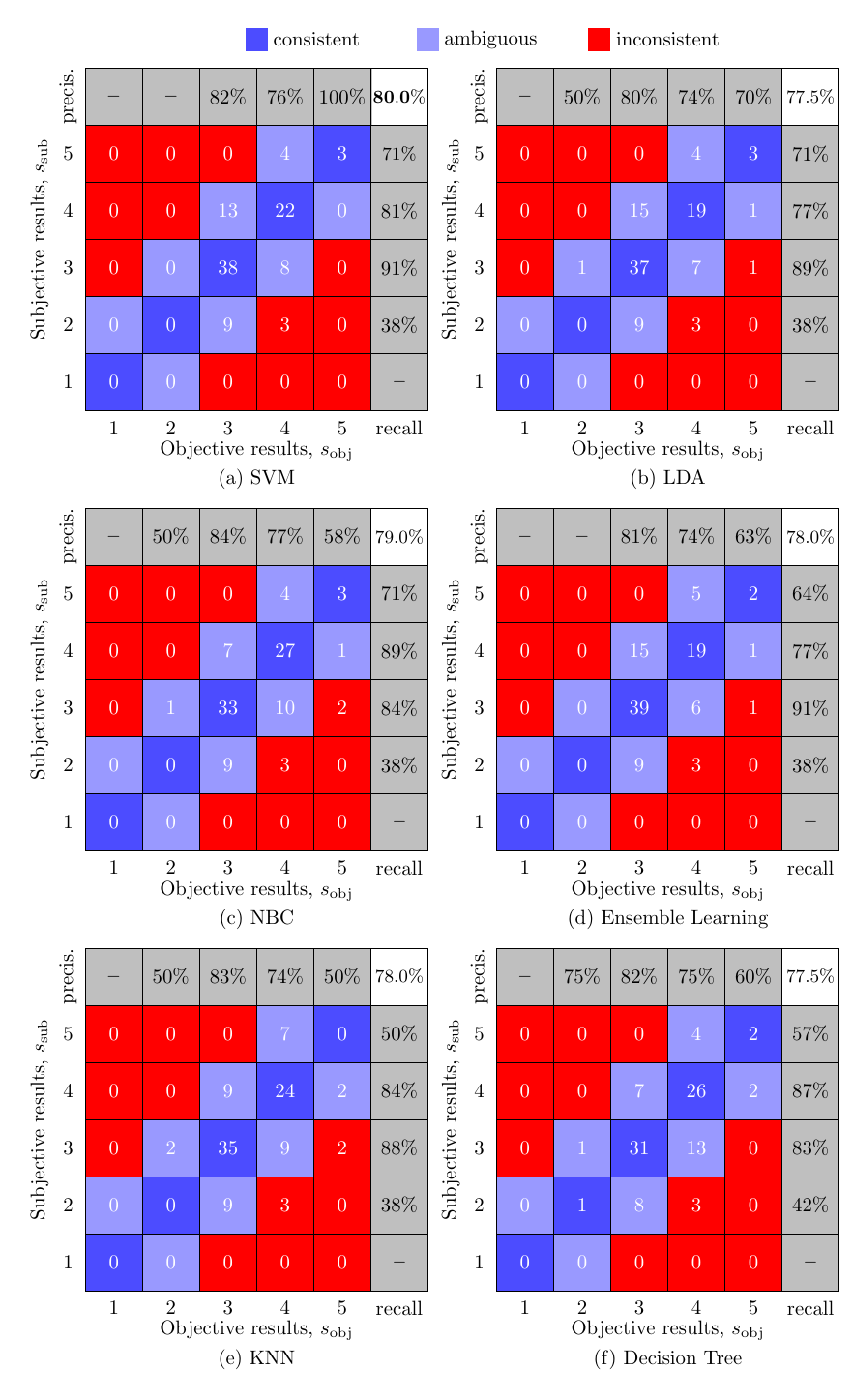}
    \caption{The weighted confusion matrix and performance in subjective-objective consistency verification. (a) SVM (b) LDA (c) NBC (d) Ensemble Classifier (e) KNN (f) Decision Tree}
    \label{fig: Sub_Obj_Confusion}
\end{figure}

We use the driving common factors and ground truth $(s_\mathrm{sub})$ of driving styles as a unified benchmark and develop $6$ driving style identifiers using five-fold cross-validation.
To illustrate the performance of the six driving style classifiers, we compare the ground truth and objective ($s_{\mathrm{obj}}$) aggressive levels of $100$ drivers generated by the classifiers, as shown in Fig. \ref{fig: Sub_Obj_Confusion}.
The differences between vertical and horizontal coordinates indicate subjective-objective inconsistency and a small difference indicates a high subjective-objective consistency. For example, when the difference between horizontal and vertical coordinates for elements on the matrix diagonal equals zero, it indicates complete consistency between subjective and objective evaluations. For convenient understanding, we named the results according to their coordinate differences as follows.
\begin{equation}
\begin{cases}
\mathrm{consistent} ,&|s_{\mathrm{obj}}-s_{\mathrm{sub}}|=0\\ \mathrm{ambiguous} ,&|s_{\mathrm{obj}}-s_{\mathrm{sub}}|=1\\ \mathrm{inconsistent} ,&|s_{\mathrm{obj}}-s_{\mathrm{sub}}|\geq2
\end{cases}
\end{equation}
It is intuitively reasonable since there does not exist an apparently strict and fully convincing boundary to distinguish two adjacent subjective/objective driving styles. Further, to consider the contribution of the subjective-objective difference to the consistency performance, we assign a consistency weight of $1.0$ to consistent results and a consistency weight of $0$ to inconsistent results. For ambiguous results, the objective results are very close to the subjective results, so we assign them a consistency weight of $0.5$.  Then, we compute the consistency, precision, and recall according to the consistency-weighted confusion matrix. 

The subjective-objective consistency verification results of the six classifiers are shown in Fig.~\ref{fig: Sub_Obj_Confusion}. Because the driving style labels obtained in Section V-A does not encompass level 1, thereby excluding it from the scope of our result discussion. All classifiers exhibit better performance on the levels 3 and 4 compared to other categories, indicating the levels 3 and 4 are easier to recognize.
Among all the classifiers, the SVM achieves the best overall performance with a general consistency of $80$\si{\%}. The SVM achieves the highest precision of $100\%$ for aggressive level 5 and a relative high precison for level 4, indicating that SVM possesses a strong capability in recognizing aggressive drivers.
The NBC demonstrates high precision and recall in the levels 3 and 4 but exhibits poorer performance in the level 5, thereby rendering it less effective than the SVM.

The above analysis illustrates that, with the help of ground truth, we can comprehensively analyze the performance of various driving style algorithms. The 100-DrivingStyle dataset serves as a unified benchmark that facilitates the evaluation of various methods, enabling the formulation of consistent conclusions and fostering innovation in the field of driving style research.



\section{Conclusion and Future Works}
Using an onboard data acquisition system, we collected 2200 km of natural driving behavior and 3000 minutes of video records. Our main contribution is a large-scale driving style dataset that aims to fill in the data gap of missing subjective driving style perception of drivers and experts in the existing natural driving dataset. Furthermore, 100 drivers with diverse styles were recruited to sequentially drive the same car on the same route, which ensures that the dataset is not affected by vehicle types and traffic routes. Moreover, the average driving time per driver reaches about 30 minutes, which ensures data sufficiency for driving style research. All this distinguishes the data set from other natural driving datasets.  Therefore, the 100-DrivingStyle dataset can facilitate the research of driving style recognition, especially the generalization of driving style recognition algorithms in multiple scenarios, the analysis of driving style in the long-term and the short term, etc. Although the 100-DrivingStyle dataset is originally designed for driving style recognition, we also hope to boost research on driver models, driver recognition, driving intention recognition, and other topics that rely on long-term naturalistic driving data.










\bibliographystyle{IEEEtran}
\bibliography{IVrefs}

\end{document}